\documentclass[journal]{IEEEtran}

\usepackage{amsmath,amsfonts}
\usepackage{algpseudocode}
\usepackage{algorithmicx,algorithm}
\usepackage{hyperref}
\usepackage{array}
\usepackage[caption=false,font=normalsize,labelfont=sf,textfont=sf]{subfig}
\usepackage{textcomp}
\usepackage{stfloats}
\usepackage{url}
\usepackage{verbatim}
\usepackage{graphicx}
\usepackage{cite}

\usepackage[square,sort&compress,numbers]{natbib}
\usepackage{amsmath}
\usepackage{amsfonts} 
\usepackage{amssymb} 
\usepackage{multirow}
\usepackage{threeparttable}
\usepackage{booktabs}

\usepackage{bbm, dsfont}
\usepackage{bbding}
\usepackage{ifsym}

\graphicspath{{figs/}}
\graphicspath{{bio/}}

\usepackage{color, xcolor}

\begin{document}
\title{MarsSeg: Mars Surface Semantic Segmentation with Multi-level Extractor and Connector}

\author{
Junbo~Li,~Keyan~Chen,~Gengju~Tian,~Lu~Li$^\star$~and~Zhenwei~Shi
\\
\vspace{4pt}
Beihang University
}

\maketitle

\setlength{\abovedisplayskip}{6pt}
\setlength{\belowdisplayskip}{2pt}

\begin{abstract}

The segmentation and interpretation of the Martian surface play a pivotal role in Mars exploration, providing essential data for the trajectory planning and obstacle avoidance of rovers. However, the complex topography, similar surface features, and the lack of extensive annotated data pose significant challenges to the high-precision semantic segmentation of the Martian surface.
To address these challenges, we propose a novel encoder-decoder based Mars segmentation network, termed MarsSeg. Specifically, we employ an encoder-decoder structure with a minimized number of down-sampling layers to preserve local details. To facilitate a high-level semantic understanding across the shadow multi-level feature maps, we introduce a feature enhancement connection layer situated between the encoder and decoder. This layer incorporates Mini Atrous Spatial Pyramid Pooling (Mini-ASPP), Polarized Self-Attention (PSA), and Strip Pyramid Pooling Module (SPPM).
The Mini-ASPP and PSA are specifically designed for shadow feature enhancement, thereby enabling the expression of local details and small objects. Conversely, the SPPM is employed for deep feature enhancement, facilitating the extraction of high-level semantic category-related information. Experimental results derived from the Mars-Seg and AI4Mars datasets substantiate that the proposed MarsSeg outperforms other state-of-the-art methods in segmentation performance, validating the efficacy of each proposed component.

\end{abstract}


\IEEEpeerreviewmaketitle
\section{Introduction}

\IEEEPARstart {T}he exploration of extraterrestrial realms, particularly outer space, carries profound implications for the future progression of humankind \cite{li2021china, delatte2019segmentation}.
Mars, Earth's nearest planetary neighbor and a potential habitat for life, has become a principal target for exploration by numerous nations. The automated segmentation, recognition, and comprehension of Martian surface features represent critical and integral stages in Mars exploration missions \cite{meng2021high, chen2024rsmamba, chen2021building}. 
These processes are foundational for efficient trajectory planning, obstacle avoidance, and asset positioning. The accuracy of surface segmentation significantly impacts the success rate of tasks associated with Mars exploration \cite{liu2022mars, zhu2023design}.

In recent years, the rapid evolution of remote sensing technology has markedly enhanced the accuracy of semantic segmentation, with significant implications for diverse fields such as urban planning and resource exploration \cite{sabins2020remote, chen2024rsprompter, li2021geographical}. Contemporary methodologies predominantly concentrate on earth observation tasks, evolving from the stages of feature extraction and classifier construction to the adoption of comprehensive end-to-end deep learning techniques \cite{chen2022resolution}.
Langkvist et al. \cite{classsat2016langkvist} introduced a Convolutional Neural Network (CNN)-based architecture for remote sensing image segmentation and object recognition. AD-LinkNet \cite{2019Towards} utilized an encoder-decoder structure, incorporating dilated convolution and channel attention mechanisms to augment image semantic segmentation precision.
The growing prevalence of attention mechanisms and foundation models has led to significant advancements in segmentation methods \cite{wang2024rsbuilding, li2022geographical, chen2023ovarnet}. These methods not only deliver superior accuracy but also demonstrate enhanced generalization capabilities across diverse scenarios. RSPrompter\cite{chen2024rsprompter} introduces a prompt learning method to generate appropriate prompts for the Segment Anything Model (SAM) \cite{kirillov2023segment}, thereby achieving high-precision automatic instance segmentation.
RSBuilding \cite{wang2024rsbuilding} improves building segmentation and change detection accuracy through federated training.

Segmentation algorithms specifically tailored for the Martian surface are in their embryonic stages and warrant further exploration. The segmentation process for Martian topography diverges markedly from that of terrestrial objects on Earth, owing to its distinctive, non-structural attributes. These complexities significantly impact the accurate delineation and comprehension of surface objects. The primary challenges are as follows: i) Minimal inter-class differences: The Martian landscape, bereft of flora, water, and essential life elements, exhibits limited diversity in its landforms. This results in less distinguishable boundaries and unique features between various landforms, presenting a considerable challenge for segmentation algorithms striving to accurately delineate Martian surface features. ii) Imbalanced categories and scale variations: Martian topography poses unique challenges for semantic segmentation due to the imbalanced categories of ground objects and substantial variations in their scales. The tasks of extracting distinguishing features between different categories and identifying similarities within a category have become increasingly intricate. For instance, the AI4Mars dataset \cite{ai4mars2021Swan}, which includes only four categories, has the ``big rock" category constituting a mere 2\% proportion. iii) Limited availability of cross-scenario samples: The inherent difficulties in procuring images from the Martian surface result in a dearth of available datasets. Currently, only a handful of datasets exist, and their sample sizes are relatively small. The extraction of cross-scenario, generalizable features from this limited pool of supervised information poses a significant challenge.

In this paper, we introduce a novel encoder-decoder-based network, MarsSeg, specifically tailored for Martian landscape segmentation. MarsSeg aims to address the challenges posed by the unstructured characteristics of Martian terrain. Specifically, we employ an encoder-decoder architecture with a reduced number of down-sampling layers, thereby enabling the preservation of fine-grained local details. To further enhance the extraction of high-level semantic features from the shadow multi-level feature maps, we propose the incorporation of three key components in our network: Mini Atrous Spatial Pyramid Pooling (Mini-ASPP), Polarized Self-Attention (PSA), and Strip Pyramid Pooling Module (SPPM). These components facilitate better feature representation and contribute to improved segmentation performance in terms of scale variations and minimal inter-class differences. Furthermore, we introduce focal-dice combined loss function with adaptive weight to optimize the network during training, which is tailored to the class-imbalanced requirements of Martian landscape segmentation, ensuring accurate and robust results. 

\vspace{6pt}
The primary contributions of the paper can be summarized as follows:

i) We propose an innovative encoder-decoder-based network architecture specifically tailored for the challenging task of unstructured Martian terrain segmentation. Our approach can effectively address the challenges posed by large-scale variations and minor inter-class differences.

ii) Specifically, we introduce feature enhancement connection layers, including shadow and deep feature connectors, between the encoder and decoder to aggregate high-level semantic features. 
The shadow feature connector comprises a stacked layer of Mini-ASPP and PSA modules to capture local details. Conversely, the deep SPPM connector is proposed to gather category-related semantic features.

iii) We extensively evaluate our method on the Mars-Seg and AI4Mars datasets. The experimental results demonstrate state-of-the-art performance in terms of objective indicators and segmentation visualization.

\vspace{6pt}

The remainder of this paper is organized as follows: Sec. II offers a comprehensive review of the relevant literature. Sec. III provides an in-depth exploration of our proposed model, with a particular emphasis on its innovative components. Sec. IV presents both quantitative and qualitative results, utilizing the Mars-Seg and AI4MARS datasets. Finally, Sec. V encapsulates the key insights of this paper.
\section{Related Works}

\subsection{Semantic Segmentation}

The primary aim of semantic segmentation is to classify each pixel within an image, thereby generating a semantic class label map the same size as the input image \cite{hao2020brief}. Unlike other perception tasks, semantic segmentation yields fine-grained, pixel-level interpretation results \cite{zou2023object, chen2023continuous}. Presently, image semantic segmentation methodologies can be broadly bifurcated into traditional methods and deep learning approaches.
Traditional machine learning methods underscore the importance of feature engineering, which encompasses both feature extraction and classifier construction. However, these traditional methodologies often necessitate manual hyperparameter adjustments tailored to specific scenarios or data sets. As a result, the segmentation process becomes complex, inefficient, and unsuitable for widespread deployment and practical applications \cite{li2018survey, minaee2021image}.

Conversely, deep learning methods have gained prominence in semantic segmentation due to their inherent ability to autonomously learn pertinent features from data \cite{ghosh2019understanding}. Leveraging robust feature extraction capabilities, deep learning-based methods achieve exceptional segmentation accuracy and demonstrate cross-scenario generalization. Early semantic segmentation models were primarily focused on the Fully Convolutional Network (FCN) \cite{2015Fully}, which accepts input of arbitrary size and produces output of corresponding size through convolution and skip layer fusion. FCN-CRF \cite{2018Roads} introduces a novel fully convolutional network and employs conditional random fields to augment the detection capabilities of small targets. Moreover, FCN-rLSTM \cite{2017FCN} incorporates atrous convolution into the FCN and merges it with the LSTM network to improve the detection accuracy of road vehicle density.

However, FCNs \cite{2015Fully, 2017FCN} exhibit limitations in delivering multiscale information and local details. In contrast, encoder-decoder structures, as exemplified by the UNet \cite{ronneberger2015unet}, successfully mitigate these limitations and have thus emerged as the dominant approach in semantic segmentation. Typically, encoder-decoder networks consist of an encoder, which progressively reduces the scale of feature maps while capturing higher-level semantic information, and a decoder, which gradually recovers spatial details to generate the semantic mask. For instance, UNet++ \cite{zhou2018unetpp} introduces a series of nested, dense skip pathways based on the UNet to enhance the segmentation accuracy of medical images.

Recently, segmentation methods that utilize Transformer \cite{vaswani2017attention} architectures have shown superior performance compared to traditional convolutional approaches. For example, MaskFormer \cite{cheng2021maskformer} employs a transformer structure to endow features with global receptive fields and utilizes optimal transport matching for supervised training, thereby challenging the paradigm of semantic segmentation based on fully convolutional network design. Additionally, SegFormer \cite{xie2021segformer} designs a hierarchically structured Transformer encoder and employs a lightweight multilayer perception as the decoder to enhance segmentation efficiency and robustness.

In our paper, we investigate Mars landscape segmentation using an encoder-decoder architecture. Specifically, we introduce a novel feature enhancement connection layer situated between the encoder and decoder components. This layer serves to abstract critical features, including shadow details (such as edges and textures) and deep semantic information, which can address the inherent limitations of the standard encoder-decoder architecture, particularly in managing scale variations and inter-class similarities.

\subsection{Spatial Pyramid Pooling}

SPP-Net \cite{SPP2015he} introduces a novel pooling strategy, termed spatial pyramid pooling, designed to capture features at various scales, thereby enhancing the network's robustness to image deformation. Drawing inspiration from the SPP, Deeplabv2 \cite{deeplab2018Chen} incorporates the atrous spatial pyramid pooling (ASPP) module, which employs parallelized dilation convolution with varying expansion rates. This innovative module effectively expands the receptive field to encapsulate multi-scale information. RFBNet \cite{liu2018RFB} inspired by the receptive field structure inherent in the human visual system, proposes a receptive field block (RFB) module to augment the ability to distinguish features and enhance robustness. However, this model falls short of directly aggregating global effective information. To address this issue, the pyramid pooling module (PPM) \cite{pspn2017zhao} adopts the concept of global average pooling, dividing a branch into multi-scale sub-branches for global average pooling, thereby bolstering the model's capacity to extract and capture features at varying levels.

Based on the ASPP module, this paper introduces the Mini-ASPP, characterized by a reduced number of branches, smaller convolution kernel sizes, and lower expansion rates. This lightweight module offers significant advantages over shallow feature extractor modeling and is capable of fulfilling regional multi-scale feature extraction with a reduced computational cost.

\begin{figure*}[!th]
\centering
\resizebox{0.99\linewidth}{!}{
\includegraphics[width=\linewidth]{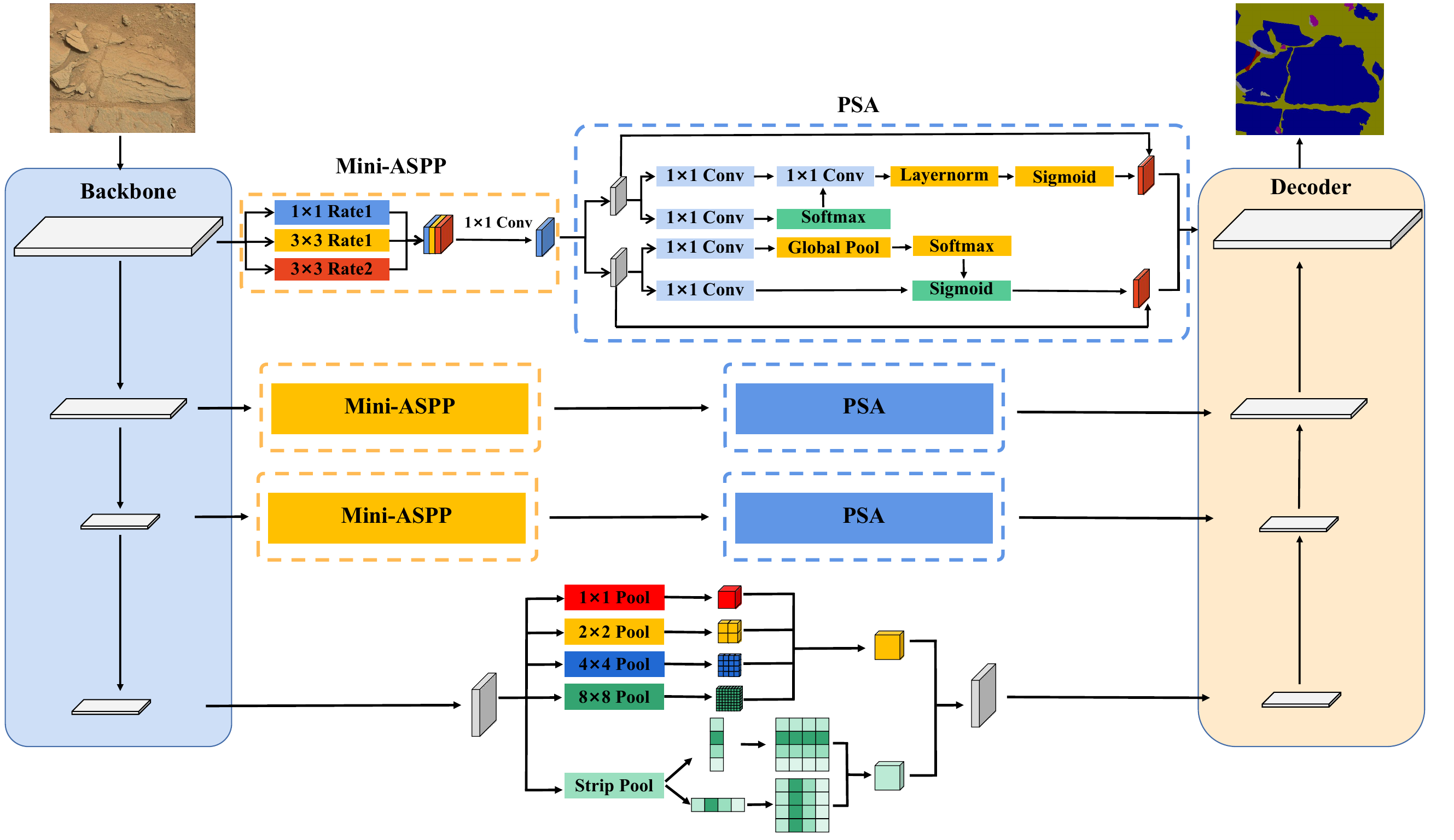}
}
\caption{
The MarsSeg framework primarily comprises three integral components: an encoder, feature enhancement connections, and a decoder. The feature enhancement connections are further subdivided into the Mini-ASPP, PSA, and SPPM. The Mini-ASPP and PSA modules collectively form the enhancement connections for shallow features, whereas the SPPM is responsible for the enhancement connections for mid-level features. These enhancement connections are effectively adept at adapting to the complex and variable Martian terrain, as well as the minimal inter-class feature differences.
}
\label{fig:model}
\end{figure*}

\subsection{Attention Mechanism}

The attention mechanism, inspired by the human vision system, empowers the model to selectively concentrate on pertinent information within images \cite{guo2022attention}. This mechanism has proven to be instrumental in semantic segmentation tasks, evolving into a fundamental module of segmentation networks due to its proficiency in efficiently constructing contextual dependencies of semantic features. The visual attention mechanism can be primarily categorized into channel attention, spatial attention, hybrid attention, Non-local, and Transformer \cite{niu2021review}.
Channel attention enables networks to discern which channel needs focus. SENet \cite{hu2019squeezeandexcitation} models the interdependence between channels to enhance the network's capacity to express features with a minimal increase in computational cost.
Spatial attention permits networks to identify the image positions that require focus. STN \cite{jaderberg2016spatial} introduces a spatial transformer capable of processing data spatially within the network, rendering the model more invariant to translation, scaling, and rotation.
Convolutional block attention module (CBAM) \cite{woo2018cbam} integrates both channel and spatial attention for feature refinement. 
Non-local networks \cite{wang2018nonlocal} compute the response of a specific point position in the network as the weighted sum of all position features in the image, thereby augmenting the model's ability to capture the dependence of long-distance features.
Transformer \cite{Vaswani2017trans} employs an entirely attention-based mechanism to accomplish natural language processing (NLP) tasks. Moreover, the vision Transformer (ViT) \cite{dosovitskiy2021vit} applies the transformer directly to the image sequence, demonstrating impressive performance in classification. Segformer \cite{xie2021segformer} amalgamates the transformer with a lightweight multi-layer perception (MLP) decoder to achieve better segmentation results.
This paper introduces the polarized self-attention mechanism (PSA) to achieve channel and spatial focus of shallow feature maps, aligning with the Martian terrain characteristics.
\begin{figure}[!t]
\centering
\resizebox{0.99\linewidth}{!}{
\includegraphics[width=\linewidth]{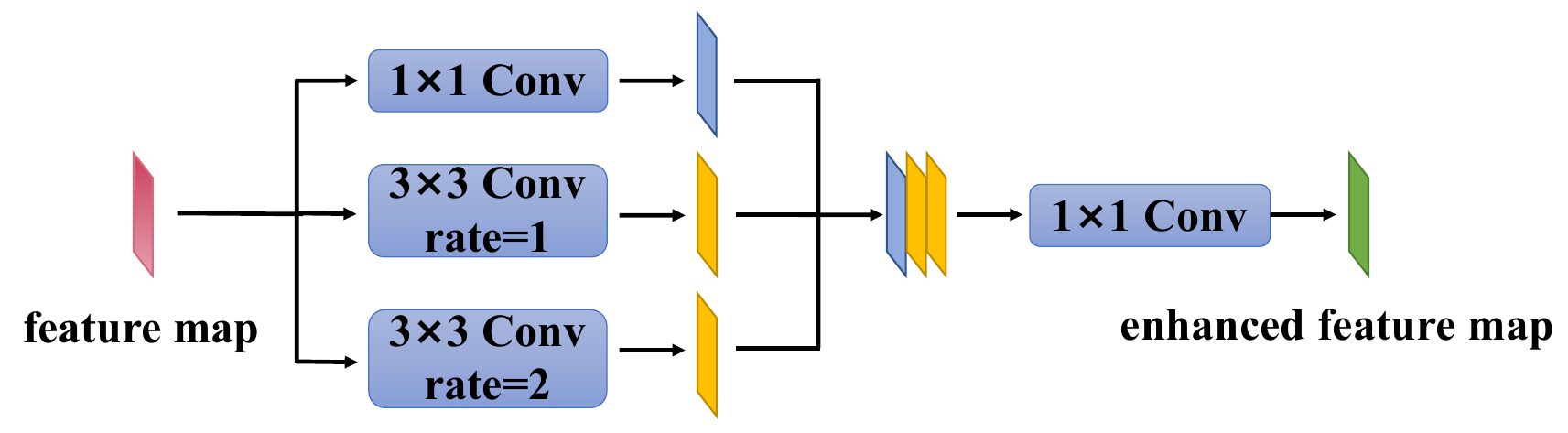}
}
\caption{
The Mini-ASPP is characterized by three distinct atrous convolution layers, each employing varying rates. This unique configuration facilitates the acquisition of local multi-scale representations efficiently. 
}
\label{fig:mini-aspp}
\end{figure}

\section{Methodology}

\subsection{Overview}

The MarsSeg architecture adheres to an encoding and decoding framework, incorporating three primary components: the encoder, the feature enhancement connection, and the decoder. This network structure is visually depicted in Fig. \ref{fig:model}. Specifically, the encoder is distinguished by three shallow feature maps of varying scales, derived from ResNet. In contrast, the decoder is assembled from multiple layers of up-sampling and convolution. The feature enhancement connection plays a crucial role, amalgamating three elements: Mini-ASPP, PSA, and SPPM.
Mini-ASPP and PSA form the enhanced connection for shallow features, while SPPM facilitates the enhanced connection for mid-level features. This enhanced connection module skillfully accommodates the intricate and dynamic characteristics of Martian terrain, effectively mitigating the limited feature distinctions between classes. The process can be elucidated as follows,
\begin{align} 
\begin{split} 
M &= \Phi_{\text{decoder}} \circ \Phi_{\text{enhancer}} \circ \Phi_{\text{encoder}} (I) \\
 \Phi_{\text{enhancer}}^i &= \begin{cases}
 \Phi_{\text{psa}} \circ \Phi_{\text{mini-aspp}}, & i<3\\
 \Phi_{\text{sppm}}, &i=3\\
 \end{cases}
\label{eq:overall} 
\end{split} 
\end{align}
where the Martian surface image, $I \in \mathbb{R}^{H \times W \times 3}$, sequentially undergoes a processing pipeline involving an encoder, a feature enhancement connection module, and a decoder to yield the semantic segmentation result, $M \in \mathbb{R}^{H \times W \times n}$, with $n$ semantic classes. $i$ represents the hierarchical level of the encoder output features. The initial two levels undergo processing via the Mini-ASPP and PSA to derive enhanced features, whereas the final level is handled by the SPPM.

\subsection{Encoder and Decoder}

The proposed MarsSeg is a segmentation network based on an encoder-decoder structure. The encoder abstracts and down-samples layer by layer to extract high-level semantic features, while the decoder reconstructs the segmentation map of various surface categories through layer-by-layer up-sampling. Specifically, we employ ResNet50 as the encoder. To prevent the loss of small objects and edge details, we exclusively utilize shallow features extracted from the backbone, \textit{i.e.}, only three feature maps of different scales, denotes as $F^i \in \mathbb{R}^{\frac{H}{2^{(i+1)}} \times \frac{W}{2^{(i+1)}} \times c}, i \in \{1, 2, 3\}$. The decoder merges the enhanced features and reconstructs the semantic segmentation map, $M \in \mathbb{R}^{H \times W \times n}$, through two up-sampling layers (Up-sampling + Convolution + Batch Normalization).

\subsection{Shadow Feature Enhancement}

\begin{figure}[t]
\centering
\resizebox{0.99\linewidth}{!}{
\includegraphics[width=\linewidth]{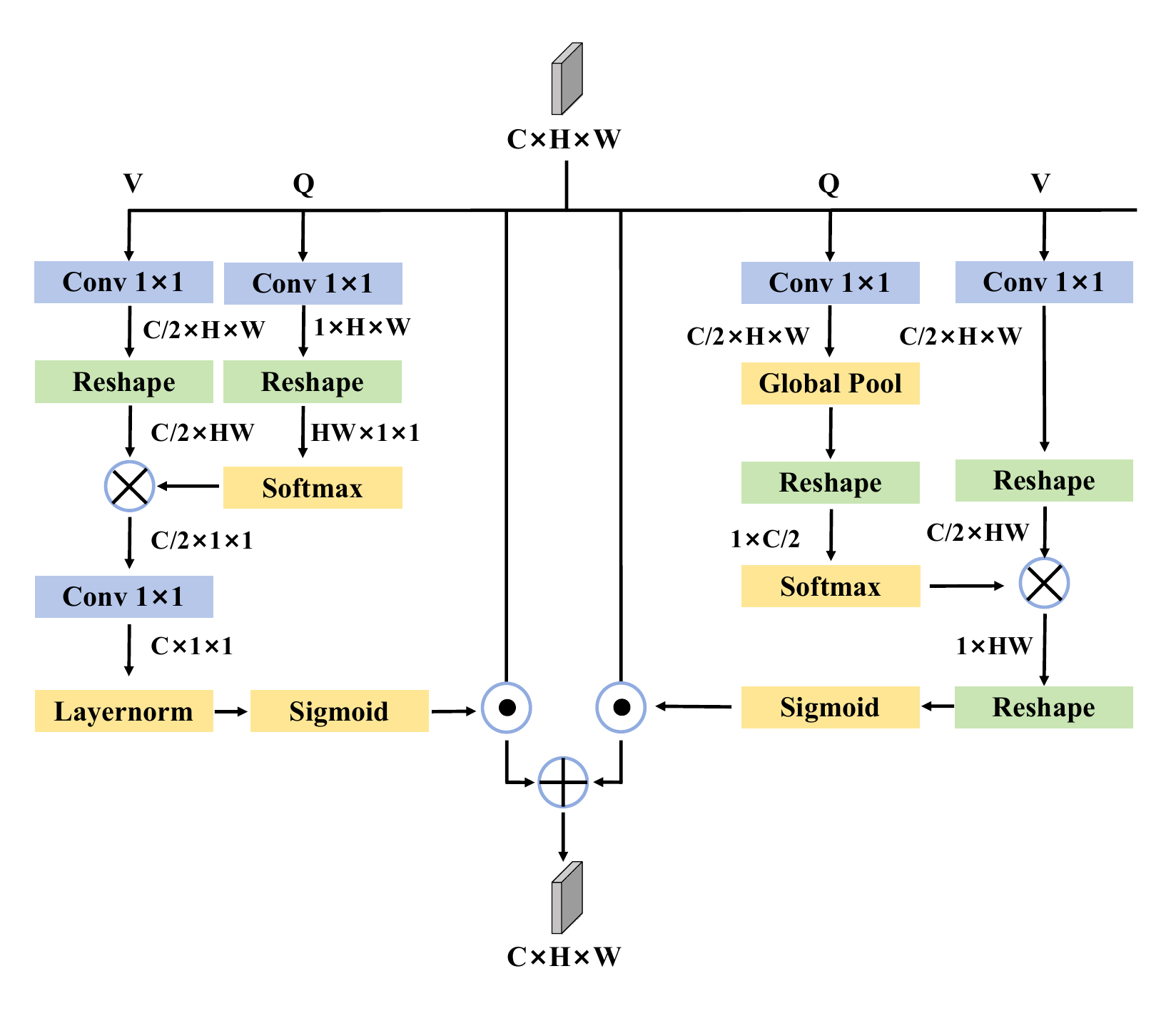}
}
\caption{
The overview of the PSA. PSA executes a lightweight self-attention process from both the channel and spatial perspectives.
}
\label{fig:psa}
\end{figure}

To optimize the segmentation effect of small targets and edge details, we introduce the Mini-ASPP and PSA modules within the skip connections of shallow features. The Mini-ASPP is employed to capture local detail features, while the PSA module is utilized to ascertain global dependencies. As we integrate feature enhancement modules into the skip connections of shallow features, \textit{i.e.}, high-resolution feature maps, computational efficiency becomes a crucial consideration.
The original Atrous Spatial Pyramid Pooling (ASPP) operates on high-level semantic (low-resolution) feature maps, where computational efficiency is not a primary concern. It employs four parallel dilated convolution kernels of varying sizes to capture multi-scale information, often using larger kernel sizes. However, if this module is applied to high-resolution feature maps, it would significantly impact computational efficiency.
Therefore, we propose a decoupling solution: the Mini-ASPP is used to efficiently establish local features, and the PSA is used to construct global dependencies. This approach enhances the feature response of small targets and edge details in high-level semantic feature maps efficiently.

The Mini-ASPP is an optimized version of the original ASPP, designed to enhance the segmentation of detailed features while reducing computational demands. This module comprises three dilated convolution layers, each with distinct kernel sizes and dilation rates, and an information aggregation layer. Specifically, the dilated convolution layers have kernel sizes of $1 \times1$, $3 \times3$, and $3 \times3$, with corresponding dilation rates of 1, 1, and 2, respectively. The information aggregation layer, on the other hand, consolidates the feature maps extracted by the three convolution layers via a $1 \times1$ convolution operation. The architectural schematic of the Mini-ASPP module is depicted in Fig. \ref{fig:mini-aspp}. The mathematical representation of this module can be expressed as follows,
\begin{align} 
\begin{split} 
F_1^i &= \Phi_{\text{dilate-conv}} (F_i, 1, 1) \\
F_2^i &= \Phi_{\text{dilate-conv}} (F_i, 3, 1) \\
F_3^i &= \Phi_{\text{dilate-conv}} (F_i, 3, 2) \\
F_i &= \Phi_{\text{cat}}([F_1^i, F_2^i, F_3^i]) \\
F_i &= \Phi_{\text{channel-reduce}} (F_i) \\
\label{eq:overall} 
\end{split} 
\end{align}
where $\Phi_{\text{dilate-conv}}(F, 3, 1)$ denotes the dilated convolution with kernel size of 3, and dilation of 1. $i$ represents the different hierarchical levels of feature maps. $\Phi_{\text{cat}}$ represents the concatenation operation along the channel dimension. $\Phi_{\text{channel-reduce}}$ represents a $1 \times1$ convolution operation employed to decrease the channel dimension.

\begin{figure}[!t]
\centering
\resizebox{0.99\linewidth}{!}{
\includegraphics[width=\linewidth]{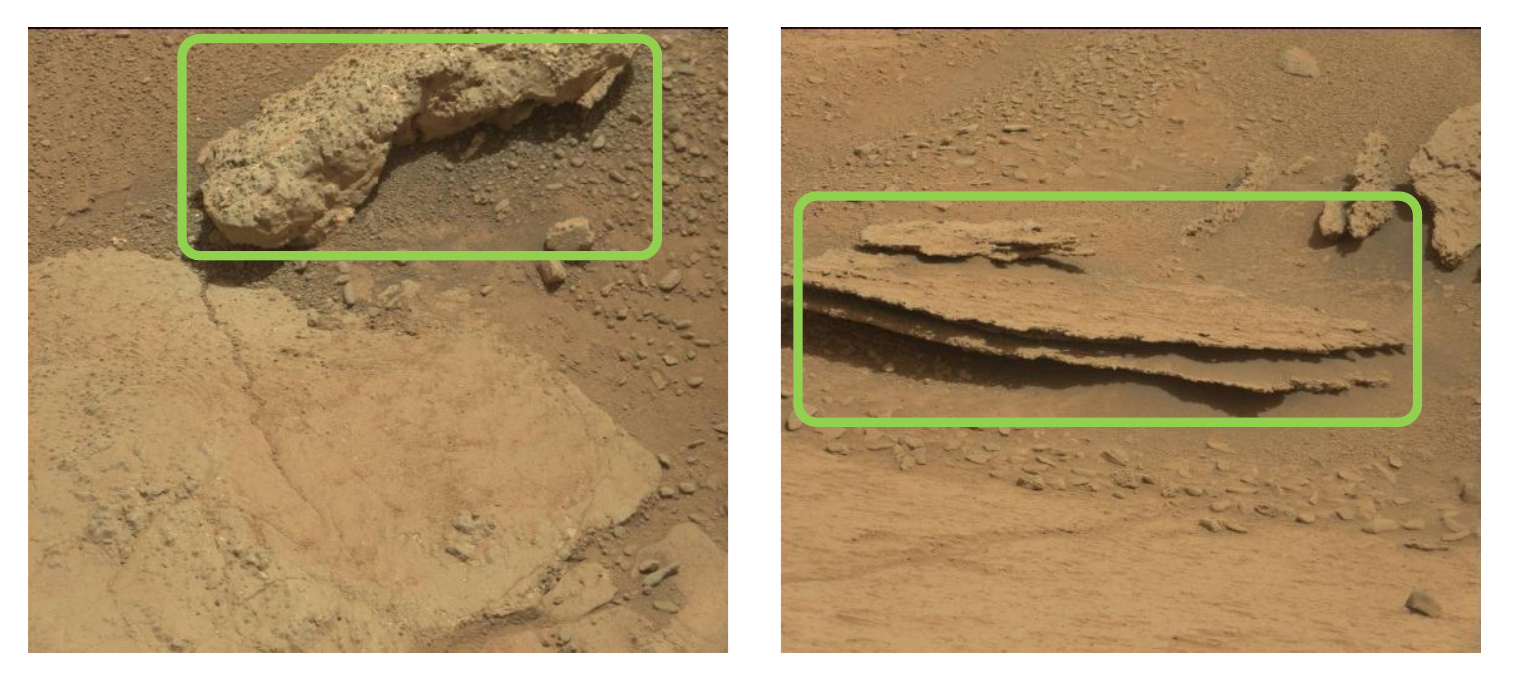}
}
\caption{Some representative examples of Mars' landforms.
}
\label{fig:vis_case}
\vspace{-8pt}
\end{figure}

While the Mini-ASPP emphasizes the expression of local details and small targets, the task of Martian surface segmentation necessitates a broader receptive field to provide semantic information that can bridge the minor inter-class feature differences and enhance the accuracy of category parsing. To address this, we introduce a polarized attention mechanism designed to extract global feature dependencies from both spatial and channel perspectives. This mechanism facilitates the smooth expression of category-discriminative features and suppresses irrelevant interference. More specifically, We proposed the Polarized Self-Attention (PSA) mechanism to achieve global attention by amalgamating channel self-attention and spatial self-attention branches. PSA employs $1 \times1$ convolution in each branch to mitigate computational complexity by reducing channels and conducting attention on channel and space, respectively. The structural diagram of the PSA is depicted in Fig. \ref{fig:psa}.

\subsection{Deep Feature Enhancement}

\begin{figure}[!t]
\centering
\resizebox{0.99\linewidth}{!}{
\includegraphics[width=\linewidth]{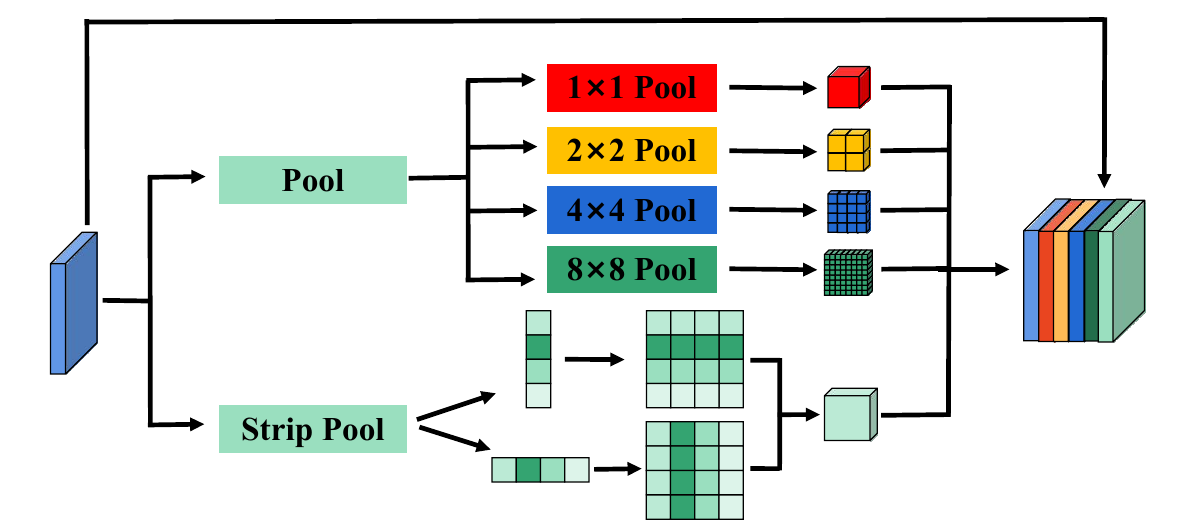}
}
\caption{
The overview of the strip pooling pyramid module (SPPM).
}
\label{fig:sppm}
\vspace{-8pt}
\end{figure}

Given that deep features tend to lose detailed information such as edges, our focus has not been on enhancing this detail within deep features. Instead, we aspire for these deep features to primarily express high-level semantic features, such as categories. Consequently, we introduce the Strip Pyramid Pooling Module (SPPM), situated between the encoder and decoder, to extract semantic information apt for the representation of Martian surface target features.
The SPPM operates concurrently with multiple distinct pooling kernels, specifically incorporating global adaptive pooling kernels pooled to dimensions of $1 \times1 $, $2\times2 $, $4 \times 4 $, $8 \times8 $, along with $1 \times W$  and $1 \times H$  stripe pooling kernels. 
The former is designed to adapt to multi-scale terrain, while the latter is tailored to accommodate stripe-shaped terrain, as depicted in Fig. \ref{fig:vis_case}. The specific structure of SPPM is illustrated in Fig. \ref{fig:sppm}.

\subsection{Loss Function}

In this paper, we employ Focal loss \cite{lin2017focal} and Dice loss \cite{sudre2017generalised} as the ultimate optimization objectives. The Focal loss assigns weights to each sample based on the complexity of sample learning, while the Dice loss mitigates the issue of significant imbalance between positive and negative samples. To accelerate the convergence rate of network training and enhance the segmentation effect, we incorporate deep supervision at each layer of the decoder to ensure semantic alignment.
Additionally, to address the problem of extreme imbalance in Martian surface category samples, we introduce an affinity loss, which assigns adaptive weights to each category. Specifically, we utilize the Intersection over Union (IoU) of each category after each validation as a weight prior, denoted as$w_k = 1 / (\text{IoU}_k + \alpha) $, $\sum w_k = 1$. $w_k$ represents the weight of the $k$th class. $\alpha$ is an offset used to mitigate the excessive weight resulting from extreme situations. The overall loss is formulated as follows,
\begin{align} 
\begin{split} 
\mathcal{L} &= \frac{1}{N} \sum_i^N (\mathcal{L}_{\text{focal}}^i + \mathcal{L}_{\text{dice}}^i) \\
\label{eq:overall} 
\end{split} 
\end{align}
where $i$ denotes the loss obtained from the $i$th feature layer, and $N$ represents the total number of layers, with $N$ equating to 3. Apart from the final layer, a convolution layer with shared parameters is utilized in conjunction with an upsampling layer to generate the category semantic probability map for each intermediate layer.
\section{Experimental Results and Analyses}

In order to evaluate the effectiveness of the proposed model, this section conducts a comparative analysis of its performance against a range of state-of-the-art deep-learning semantic segmentation models, including UNet \cite{ronneberger2015unet}, Deeplabv3+ \cite{chen2018encoderdecoder}, PSPNet \cite{pspn2017zhao}, CPNet \cite{context2020Yu}, OCRnet \cite{yuan2021segmentation}, SAW-SAN \cite{SADG2022Peng}, Cov-DA \cite{li2022stepwise}, Segformer \cite{xie2021segformer}, Seaformer \cite{wan2023seaformer}, and Lawin \cite{yan2023lawin}.

\subsection{Experimental Dataset and Settings}
\label{sec:dataset}

Experiments were conducted on two publicly accessible datasets for Mars surface segmentation: Mars-Seg \cite{Stepwise2022Li} and AI4Mars \cite{ai4mars2021Swan}.

\vspace{3pt}
\noindent \textbf{Mars-Seg} \cite{Stepwise2022Li}: 
The Mars-Seg dataset, encompassing high-resolution imagery of diverse Martian landscapes, is bifurcated into two distinct subsets: the MSL-Seg dataset, which is composed of RGB images, and the MER-Seg dataset, characterized by grayscale images. The MSL-Seg dataset comprises a total of 4155 RGB images, each with a resolution of $560 \times 500$, while the MER-Seg dataset includes 1024 grayscale images, each of size $ 1024 \times 1024$. The pixel of each image was divided into 9 categories: Martian soil, Sands, Gravel, Bedrock, Rocks, Tracks, Shadows, Background, and Unknown. For this paper, these two datasets are randomly partitioned into training and testing sets, adhering to an $8:2$ ratio.

\vspace{3pt}
\noindent\textbf{AI4Mars} \cite{ai4mars2021Swan}:
The AI4Mars dataset, launched by NASA in 2021, comprises approximately 350,000 images captured and collated by the Curiosity, Opportunity, and Perseverance Mars rovers. Presently, a subset of AI4Mars, known as AI4Mars (MSL), has been made publicly accessible. This subset encompasses 16,064 training images and 322 testing images, representing four distinct types of Martian terrain: Soil, Bedrock, Sand, and Big Rock.

\subsection{Evaluation Protocol and Metrics}

In this paper, we align with the prevalent metrics for semantic segmentation, specifically, the Intersection over Union (IoU) for each category. Concurrently, the mean IoU across all categories is employed as a comprehensive metric. The specific computation method is delineated as follows,
\begin{align} 
\begin{split} 
\text{IoU}_{k} &= \text{TP}_{k} / (\text{TP}_{k} + \text{FP}_{k} + \text{FN}_{k}) \\
\text{mIoU} &= \frac{1}{N} \sum_k^N (\text{IoU}_{k}) \\
\label{eq:iou} 
\end{split} 
\end{align}
where $\text{TP}_{k}$, $\text{FP}_{k}$, and $\text{FN}_{k}$ denote the true positive samples, false positive samples, and false negative samples of the $k$th category, respectively.

\subsection{Implementation Details}

The method employed in this paper is underpinned by the Pytorch deep learning framework and executed on an NVIDIA GeForce RTX 3090 GPU. The training process is configured with a batch size of 8, leveraging a stochastic gradient descent (SGD) algorithm characterized by a learning rate of 0.001, a momentum of 0.9, and a weight decay of 0.0001. In relation to the Mars-Seg dataset, we utilize data augmentation strategies such as random cropping, flipping, and random scaling. Conversely, for the AI4Mars dataset, data augmentation is exclusively applied to the big rock class in an effort to mitigate the challenge posed by its sparse sample ratio.

\begin{table*}[!thbp] 
\centering
\caption{
Comparison with other methods on the Mars-Seg (MSL-Seg) test set.
}
\label{tab:marsseg_msl}
\resizebox{0.95\linewidth}{!}{
\begin{tabular}{c| *{9}{c}|c}
\toprule
Method 
& $\text{Martian soil}$ 
& $\text{Sands}$ 
& $\text{Gravel}$ 
& $\text{Bedrock}$ 
& $\text{Rocks}$ 
& $\text{Tracks}$ 
& $\text{Shadows}$ 
& $\text{Background}$ 
& $\text{Unknown}$ 
& $\text{mIoU}$
\\
\midrule
UNet \cite{ronneberger2015unet} & 
15.84 & 55.89 & 70.92 &	69.19 &	38.70 &	52.50 &	43.64 &	71.64&	21.07 &	48.82
\\
Deeplabv3+ \cite{chen2018encoderdecoder} & 
30.41 &	62.43 &	73.02 &	73.99 &	34.21 &	53.81 &	43.84 &	67.10 &	32.17 &	52.33
\\
PSPNet \cite{pspn2017zhao} &
40.75 &	72.26 &	83.82 &	81.76 &	46.82 &	64.32 &	50.04 & 74.88 & 35.26 &	61.06
\\
CPNet \cite{context2020Yu} & 
42.65 &	68.44 &	82.53 &	82.95 &	52.14 &	75.35 &	51.01 & 88.82 &	21.37 &	62.81
\\
OCRNet \cite{yuan2021segmentation} & 
38.40 &	67.66 &	78.56 &	78.31 &	46.66 &	72.17 &	51.94 &	83.22 &	34.08 &	61.22
\\
SAN-SAW \cite{SADG2022Peng} & 
39.47 &	69.08 &	76.66 &	78.23 &	48.57 &	78.49 &	52.60 &	79.26 &	35.74 &	62.01
\\
Cov-DA \cite{li2022stepwise} & 71.55 & 75.28 &	64.09 &	78.73 &	48.89 &	63.42 &	38.51 &	67.97 & 73.41 &	64.76
\\
Segformer \cite{xie2021segformer} & 34.56 &	65.95 &	76.26 &	76.67 &	44.24 &	55.22 &	50.43 &	77.01 &	42.29 &	58.07
\\
Seaformer \cite{wan2023seaformer} & 
32.85 &	66.94 &	77.13 &	75.59 &	44.84 &	59.32 &	47.64 &	79.45 &	41.07 &	58.32
\\
Lawin \cite{yan2023lawin} & 
38.73 &	66.82 &	71.64 &	72.46 &	44.48 &	63.96 &	46.49 &	79.70 &	40.94 &	58.36
\\

\midrule
MarsSeg (Ours) & 44.87 & 72.23 & 79.56 & 79.48 & 52.81 & 79.63 & 55.99 & 84.97 & 41.71 & \textbf{65.69}
\\
\bottomrule
\end{tabular}
}
\end{table*}

\begin{table*}[!thbp] 
\centering
\caption{
Comparison with other methods on the Mars-Seg (MER-Seg) test set.
}
\label{tab:marsseg_mer}
\resizebox{0.95\linewidth}{!}{
\begin{tabular}{c| *{9}{c}|c}
\toprule
Method 
& $\text{Martian soil}$ 
& $\text{Sands}$ 
& $\text{Gravel}$ 
& $\text{Bedrock}$ 
& $\text{Rocks}$ 
& $\text{Tracks}$ 
& $\text{Shadows}$ 
& $\text{Background}$ 
& $\text{Unknown}$ 
& $\text{MIOU}$
\\
\midrule
Deeplabv3+ \cite{deeplabv3p2020Yang} & 
18.32 &	33.96 &	68.25 &	53.72 &	48.96 &	0.96 &	33.04 &	86.13 &	85.84 &	47.69
\\
PSPNet \cite{pspn2017zhao} &
15.76 &	42.64 &	68.34 &	60.87 &	49.01 &	28.33 &	27.81 &	87.41 &	72.84 &	51.59
\\
CPNet \cite{context2020Yu} & 
3.36 &	35.69 &	67.31 &	53.31 &	41.30 &	13.23 &	14.83 &	92.07 &	83.62 &	44.98
\\
OCRNet \cite{yuan2021segmentation} & 
0.00 &	33.54 &	68.39 &	58.35 &	51.28 &	33.07 &	29.84 &	89.86 &	69.44 &	48.11
\\
SAN-SAW \cite{SADG2022Peng} & 
17.84 &	33.45 &	73.13 &	44.94 &	33.19 &	29.79 &	38.12 &	88.84 &	86.41 &	50.69
\\
Segformer \cite{xie2021segformer} & 
8.75 &	41.08 &	72.70 &	58.70 &	46.69 &	18.62 &	20.22 & 97.46 &	89.01 &	50.36
\\
Seaformer \cite{wan2023seaformer} & 
14.12 &	36.19 &	65.50 &	50.26 &	34.08 &	27.95 &	42.52 &	85.51 &	89.54 &	49.52
\\
Lawin \cite{yan2023lawin} & 
16.58 &	34.21 &	66.19 &	52.99 &	35.48 &	20.60 &	35.54 &	78.73 & 91.01 &	47.95
\\

\midrule
MarsSeg (Ours) & 25.20& 37.47 & 71.54 & 58.67 & 47.44 & 26.30 &	32.32 & 85.23 & 87.84 & \textbf{52.48}
\\
\bottomrule
\end{tabular}
}
\end{table*}

\subsection{Comparison with the State-of-the-Arts On Mars-Seg Dataset}

The Mars-Seg segmentation dataset is bifurcated into two subsets, specifically, MSL-Seg, comprising RGB images, and MER-Seg, encompassing grayscale images. Experiments were carried out on both subsets, with Tab. \ref{tab:marsseg_msl} and Tab. \ref{tab:marsseg_mer} delineating the comparative results between the proposed method and other state-of-the-art techniques. It can be seen from Tab. \ref{tab:marsseg_msl} that the proposed method demonstrates superior performance, with exceptional results across most of the subclasses. In comparison to the baseline Deeplabv3+, the average IoU has escalated from 52.33 to 65.69. Concurrently, our approach significantly outperforms other attention-based techniques such as Segformer, Seaformer, and Lawin. This superiority may be attributed to the fact that the methodology developed in this paper can also obtain benefits when the training data is scarce. Tab. \ref{tab:marsseg_mer} presents a comparison of the segmentation effects of grayscale images, from which it is evident that the proposed model also attained the highest segmentation accuracy. It is noteworthy that our approach can exhibit a significant performance enhancement in categories with a smaller sample proportion, such as Martian soil, owing to our specific loss function design.

\begin{figure*}[!thpb]
\centering
\resizebox{0.95\linewidth}{!}{
\includegraphics[width=\linewidth]{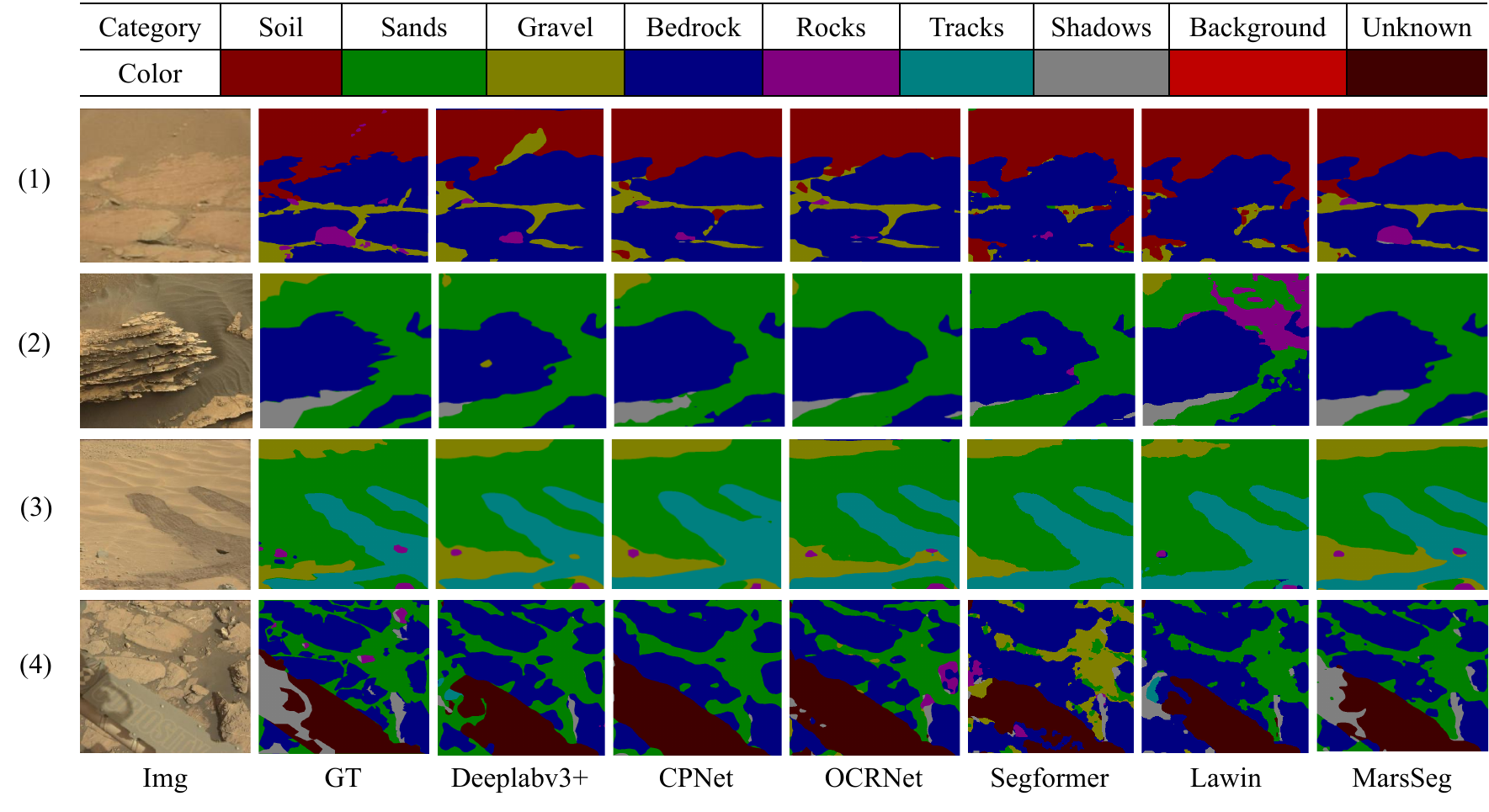}
}
\caption{
Visual comparison with other state-of-the-art methods on the Mars-Seg (MSL-Seg) test set.
}
\label{fig:marsseg_msl}
\end{figure*}

\begin{figure*}[!thpb]
\centering
\resizebox{0.95\linewidth}{!}{
\includegraphics[width=\linewidth]{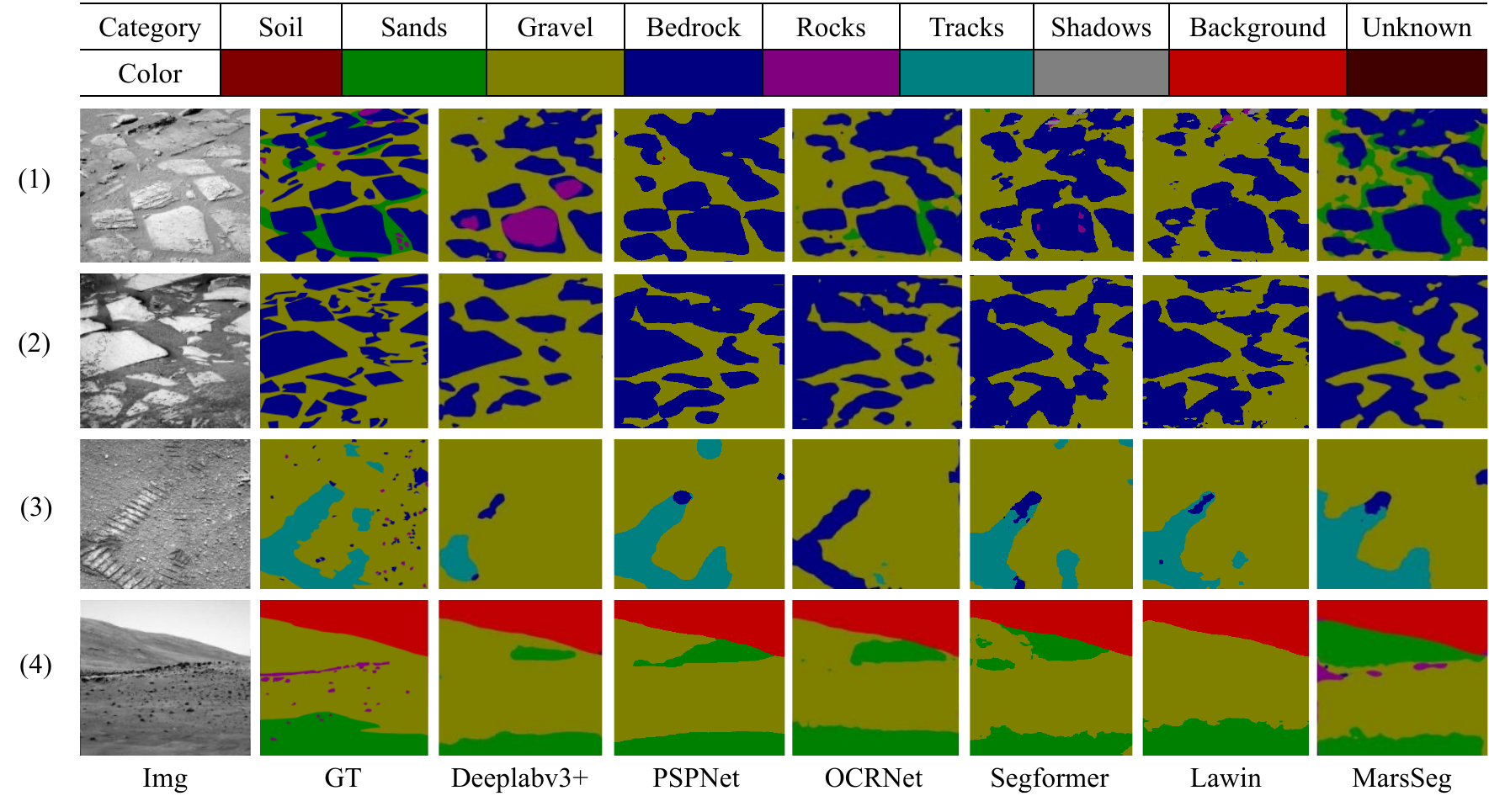}
}
\caption{
Visual comparison with other state-of-the-art methods on the Mars-Seg (MER-Seg) test set.
}
\label{fig:marsseg_mer}
\end{figure*}

Fig. \ref{fig:marsseg_msl} and Fig. \ref{fig:marsseg_mer} respectively offer a visual comparison between the proposed method and other state-of-the-art methods on the two subsets. As evident from Fig. \ref{fig:marsseg_msl}, Segformer is deficient in handling small targets and blurred boundaries, leading to numerous incorrect segmentations. The proposed method not only improves the segmentation ability of small targets and blurred boundaries, but also distinctly differentiates similar categories, such as bedrock and rocks, as illustrated in Fig. \ref{fig:marsseg_msl} (1) with rocks embedded in the bedrock. Identifying terrain and terrain boundaries from shadow areas poses a significant challenge. As discernible from the figure, the proposed method can exhibit higher accuracy in these cases, owing to the multi-scale capture of terrain features. Fig. \ref{fig:marsseg_mer} depicts the segmentation results on grayscale images. As evident from the figure, even in the absence of color information, MarsSeg can still hold a significant advantage in identifying various terrains, with fewer misclassified objects.

\begin{table}[!t]
\centering
\caption{Comparison with other methods on the AI4Mars test set.}
\begin{tabular}{c| *{4}{c}|c}
\toprule
Method 
& $\text{Soil}$ 
& $\text{Bedrock}$ 
& $\text{Sand}$ 
& $\text{Big Rock}$ 
& $\text{mIoU}$
\\
\midrule
Deeplabv3+ \cite{chen2018encoderdecoder} & 
92.88 &	93.51 &	85.26 &	46.28 &	79.48
\\
PSPNet \cite{pspn2017zhao} &
92.27 &	92.92 &	85.53 &	43.05 &	78.44
\\
CPNet \cite{context2020Yu} &
93.33 &	93.66 &	85.66 &	47.36 &	80.01
\\
OCRNet \cite{yuan2021segmentation} &
93.09 &	93.49 &	85.64 &	46.35 &	79.64
\\
\midrule
MarsSeg (Ours) & 93.08 & 93.81 & 86.17 & 50.50 & \textbf{80.89}
\\
\bottomrule
\end{tabular}
\label{tab:ai4mars}
\end{table}

\begin{figure*}[!thpb]
\centering
\resizebox{0.85\linewidth}{!}{
\includegraphics[width=\linewidth]{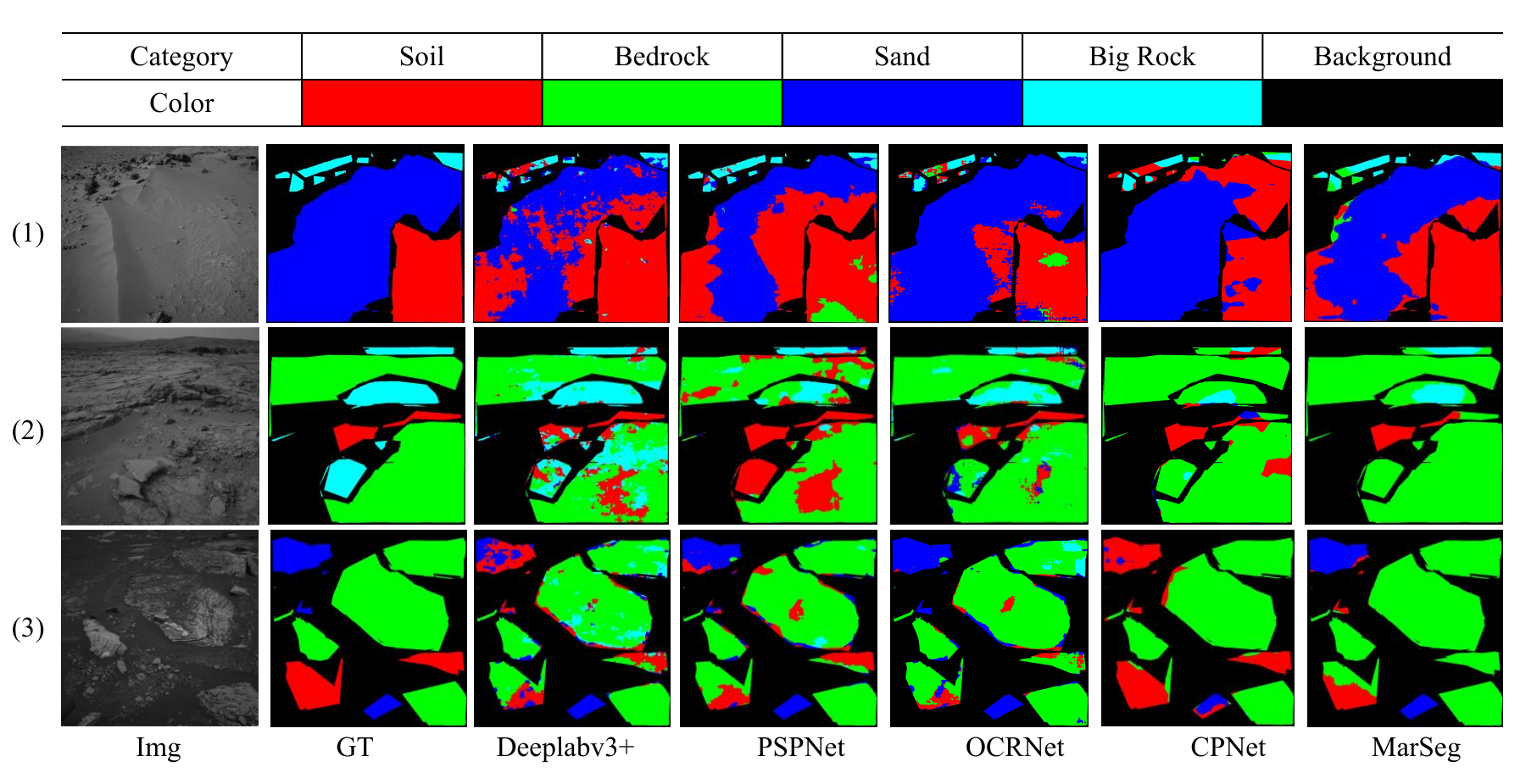}
}
\caption{
Visual comparison with other state-of-the-art methods on the AI4Mars test set.
}
\label{fig:ai4mars}
\end{figure*}

\subsection{Comparison with the State-of-the-Arts On AI4Mars Dataset}

Tab. \ref{tab:ai4mars} delineates the Intersection over Union (IoU) for various categories and mean IoU metrics within the AI4Mars dataset. As seen from the table, these alternative methods can yield high performance when the quantity of training samples is substantial (as seen in the first three categories), with an IoU consistently exceeding 85. However, when the volume of training samples is minimal (less than 2\%), and severe class imbalance issues emerge, their segmentation performance leaves much to be desired.
In contrast, the proposed method significantly outperforms these other methods in the Big Rock category, where training samples are sparse, ultimately attaining the highest average IoU of 80.89. Comparative visual segmentation results, as depicted in Fig. \ref{fig:ai4mars}, reveals the robustness of the proposed method in handling large scale variations. MarsSeg is not only capable of representing large-scale terrain but also adept at segmenting small rocks, thereby mitigating the degree of confusion among different types of land features in grayscale images with extremely weak feature distinguishability.

\subsection{Ablation Studies}

\begin{table*}[!thbp] 
\centering
\caption{
Effects of different components in the proposed MarsSeg. Experiments are conducted on the Mars-Seg (MSL-Seg) test set.
}
\label{tab:ablation}
\resizebox{0.99\linewidth}{!}{
\begin{tabular}{c c c c| *{9}{c}|c}
\toprule
SPPM & PSA & Mini-ASPP & Balancing Loss
& $\text{Martian soil}$ 
& $\text{Sands}$ 
& $\text{Gravel}$ 
& $\text{Bedrock}$ 
& $\text{Rocks}$ 
& $\text{Tracks}$ 
& $\text{Shadows}$ 
& $\text{Background}$ 
& $\text{Unknown}$ 
& $\text{mIoU}$
\\
\midrule
& & & & 
30.41 & 62.43 & 73.02 & 73.99 &	34.21 &	53.81 &	43.84 &	67.10 &	32.17 &	52.33
\\
\checkmark & & & &	
42.45 &	66.38 &	75.64 &	77.62 &	48.41 &	74.79 &	51.97 &	77.99 &	38.19 &	61.54
\\
\checkmark & \checkmark & & &	
41.18 &	67.34 &	75.53 &	76.94 &	48.03 &	80.17 &	54.06 &	79.55 &	27.52 &	61.15
\\
\checkmark & \checkmark & \checkmark & &	
40.58 &	68.27 &	77.93 &	77.77 &	51.27 &	77.60 &	53.58 &	82.93 &	38.09 &	63.11
\\
\checkmark & \checkmark & \checkmark & \checkmark &	
44.87 &72.23 & 79.56 & 79.48 & 52.81 & 79.63 & 55.99 & 84.97 & 41.71 & 65.69
\\
\bottomrule
\end{tabular}
}
\end{table*}

\begin{figure*}[!thpb]
\centering
\resizebox{0.95\linewidth}{!}{
\includegraphics[width=\linewidth]{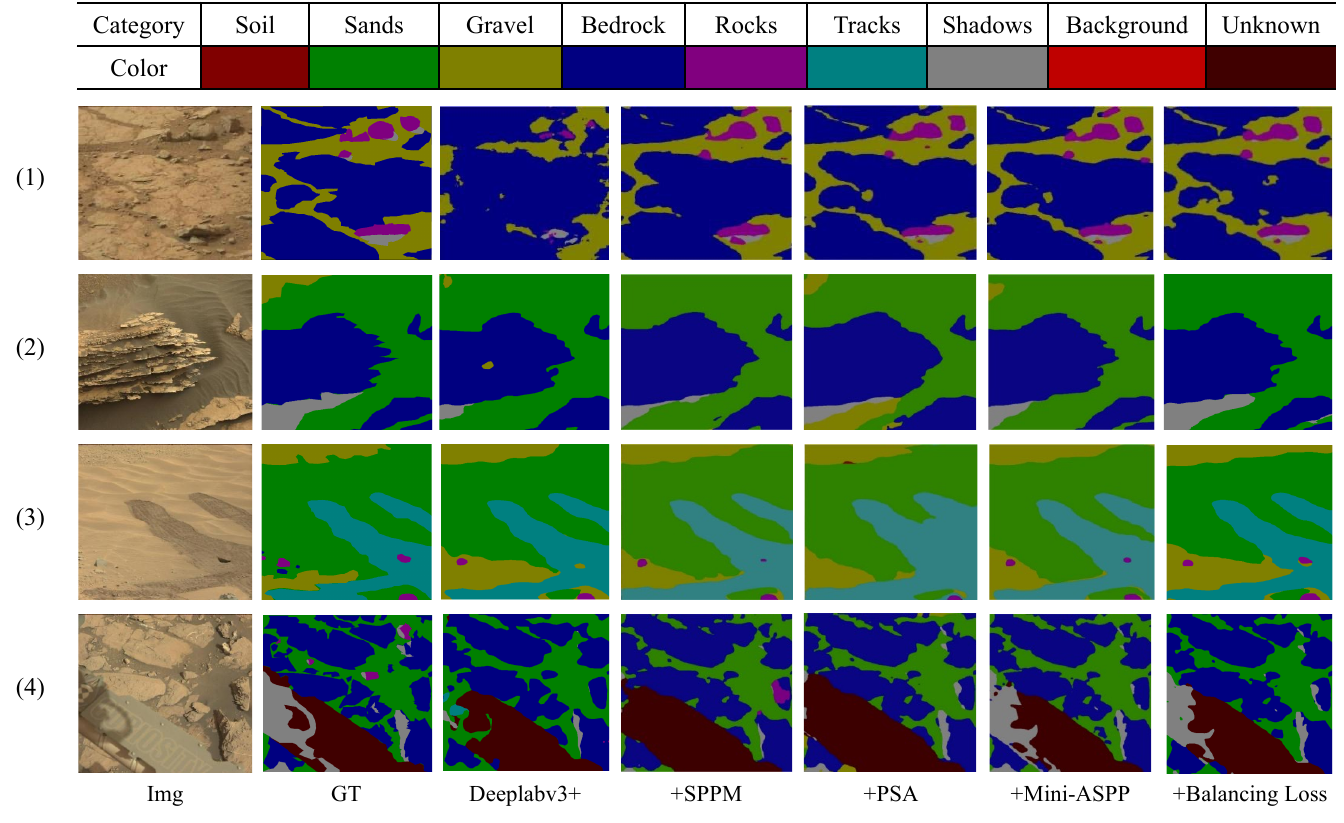}
}
\caption{
Visual comparison of different components in the proposed MarsSeg. Samples are from the Mars-Seg (MSL-Seg) test set. The components have been incorporated item by item from the left to the right.
}
\label{fig:ablation}
\end{figure*}

To substantiate the advantages of various modules, we executed ablation experiments on the Mars-Seg (MSL-Seg) dataset, the results are delineated in Tab. \ref{tab:ablation}. Several observations can be made in the table: i) In comparison to the Atrous Spatial Pyramid Pooling (ASPP) module in Deeplabv3+, the SPPM demonstrated a significant enhancement in performance, with the mIoU increasing from 52.33 to 61.54. This suggests that SPPM is more proficient at extracting the contextual information of terrestrial objects, thereby offering a more distinct classification surface of features. ii) Upon the introduction of PSA, there was a marginal decline in the model's performance, yet it exhibited a substantial improvement for the Tracks and Background categories. This implies that PSA can augment the segmentation effect of large-scale terrain. iii) The integration of Mini-ASPP further bolstered the model's performance, demonstrating a significant enhancement in the segmentation of smaller terrestrial objects, such as sands and rocks. iv) The inclusion of category balance loss culminated in the segmentation effect reaching its zenith, yielding a significant gain on categories with a lower proportion of training samples.

Fig. \ref{fig:ablation} presents the segmentation results of the corresponding methods. As seen from the figure, the original Deeplabv3+ has challenges with blurry edges, category discrimination, and small target segmentation. The incorporation of SPPM and PSA improved the accuracy of the boundaries. The precision of category discrimination was further augmented with the introduction of Mini-ASPP. The model's recognition and segmentation of small targets became more precise after the introduction of category balance loss. In conclusion, the components designed in this paper exhibit robust ability against non-structural, small inter-class differences, and blurry edges in Martian terrain segmentation.

\section{Conclusion}

In this study, we introduce MarsSeg, a Mars surface segmentation network based on the encoder-decoder structure. MarsSeg is designed to tackle the challenges posed by the unstructured nature of Martian terrain. Specifically, we diminish the quantity of downsampling layers within the encoder-decoder structure, thereby retaining fine-grained local details. To further augment the extraction of high-level semantic features from shallow multi-level feature maps, we incorporate three pivotal components into our network: Mini Atrous Spatial Pyramid Pooling (Mini-ASPP), Polarized Self-Attention (PSA), and Strip Pyramid Pooling Module (SPPM). These elements contribute to superior feature representation and enhance segmentation performance, including scale variation and minimal class differences. Additionally, we implement a loss function amalgamated with Focal-Dice, complemented by adaptive weights to optimize network training. This is specifically tailored to meet the class imbalance requirements of Martian terrain segmentation, thereby ensuring precise and robust results. Comprehensive experiments have been conducted on the Mars-Seg and AI4Mars datasets. Experimental results demonstrate that our method delivers state-of-the-art performance in terms of objective metrics and segmentation visualization.

\ifCLASSOPTIONcaptionsoff
  \newpage
\fi

\bibliographystyle{IEEEtran}
\bibliography{IEEEabrv,myreferences}

\end{document}